\documentclass[conference]{IEEEtran}
\usepackage{blindtext, graphicx}
\usepackage{subcaption}
\usepackage{amsmath}
\usepackage{ifpdf}
\usepackage{latexsym,amsfonts,amssymb,amsmath,graphicx,epsf,cite,bbm,float}
\usepackage{ifpdf,flexisym}
\usepackage{epstopdf}
\usepackage{caption}
\usepackage{subcaption}
\usepackage{comment}
\usepackage{amsthm}
\usepackage{mathtools}
\usepackage{bbm}
\usepackage{lettrine}
\usepackage{algorithm}
\usepackage{algorithmic}
\usepackage{rotating}
\usepackage{xfrac}
\usepackage[colorinlistoftodos,prependcaption,textsize=tiny]{todonotes}
\usepackage{lipsum,multicol}
\newtheorem{theorem}{Theorem}

\theoremstyle{definition}
\newtheorem{definition}{Definition}

\theoremstyle{remark}

\begin{document}
 
\title{Fairness in Supervised Learning:\\An Information Theoretic Approach}

\author{
AmirEmad Ghassami$^*$, Sajad Khodadadian$^*$, Negar Kiyavash$^{*\dagger}$\\
Departments of ECE$^*$ and ISE$^\dagger$, and Coordinated Science Laboratory$^*$,\\
University of Illinois at Urbana-Champaign, Urbana, USA.\\
\texttt{\{ghassam2,sajadk2,kiyavash\}@illinois.edu}\vspace{1mm}
}

\maketitle

\begin{abstract}

 Automated decision making systems are increasingly being used in real-world applications. In these systems for the most part, the decision rules are derived by minimizing the training error on the available historical data. Therefore, if there is a bias related to a sensitive attribute such as gender, race, religion, etc. in the data, say, due to cultural/historical discriminatory practices against a certain demographic, the system could continue discrimination in decisions by including the said bias in its decision rule. We present an information theoretic framework for designing fair predictors from data, which aim to prevent discrimination against a specified sensitive attribute in a supervised learning setting. We use equalized odds as the criterion for discrimination, which demands that the prediction should be independent of the protected attribute conditioned on the actual label. To ensure fairness and generalization simultaneously, we compress the data to an auxiliary variable, which is used for the prediction task. This auxiliary variable is chosen such that it is decontaminated from the discriminatory attribute in the sense of equalized odds. The final predictor is obtained by applying a Bayesian decision rule to the auxiliary variable.


\end{abstract}


\section{Introduction}
\label{sec:intro}

Automated decision making systems based on statistical inference and learning are increasingly common in a wide range of real-world applications such as health care, law enforcement, education, and finance. These systems are trained based on historical data, which might be biased towards certain attributes of the data points \cite{dwork2012fairness, hardt2016equality, celis2017ranking}. Hence, such data without noticing possible biases could result in discrimination, which is defined as gratuitous distinction between individuals with different sensitive attribute. These attributes include sex, race, religion, and are referred to as protected attributes in the literature.
As an example, in the US justice system, courts use features of criminals such as their age, race, sex, years being in jail, etc., to estimate their possible recidivism--future arrest. After considering these features, the court assigns a score to each in-jail individual, and decides on whether to release that person. If the score exceeds some certain limit, it will be safe to release that individual. For instance, as noted by Angwin et al. analysis \cite{angwin2016machine}, risk scores in the criminal justice system--the COMPAS risk tool--are biased negatively towards African-Americans. They showed that this risk score unjustifiably shows high risk of recidivism for African-American people compared to what it should actually be.
As another example, the authors in \cite{kay2015unequal} have studied the accuracy of gender representation in online image searches. The results indicate that for instance, in a Google image search for ``C.E.O.'', 11 percent of the depicted results are women, even though 27 percent of U.S. chef executives are women; and in a search for ``telemarketer'', 64 percent of the people depicted were female, while the occupation is evenly split between men and women.

There is an interesting connection between the problem of fairness and differential privacy \cite{dwork2008differential, dwork2006calibrating, kalantari2016optimal}. As in the differential privacy problem, one tries to hide the identity of individuals, in the fairness problem, the goal is to hide the information about the protected attribute. More details regarding this connection is presented in \cite{dwork2012fairness}.

Different criteria for assessing discrimination has been suggested in the literature. The most commonly used criterion is the so-called \textit{demographic parity}, which requires the predictor to be statistically independent from the protected attribute. That is, denoting the protected attribute and the prediction by $A$ and $\hat{Y}$, respectively, demographic parity requires the model to satisfy
\[
P(A,\hat{Y})=P(A)P(\hat{Y}).
\]
While demographic parity and its variants have been used in several works \cite{zemel2013learning, feldman2015certifying, zafar2017fairness, edwards2015censoring}, in some scenarios this criterion fails to provide fairness to all demographics \cite{dwork2012fairness}. For example, in the case of hiring an employee, where majority of the applicants are from a certain demographic, if we force the decision making system to be independent of that demographic, the system has to pick equal number of applicants from each demographic. Therefore, the system may admit a lower qualified individual from the smaller demographic to guarantee that the percentages of hired people from different demographics matches. Moreover, denoting the true label by $Y$, in most of the cases, as in the image search example, $Y$ is correlated with the protected attribute (see Figure \ref{fig:GM}). Therefore, as demographic parity forces $\hat{Y}$ to be independent of $A$, this criterion will not be satisfied for the ideal predictor $\hat{Y}=Y$.

Hardt, Price and Srebro have recently proposed \textit{equalized odds} as a new criterion of fairness \cite{hardt2016equality}. This notion demands that the predictor should be independent of the protected attribute conditioned on the actual label $Y$. Therefore, equalized odds requires the model to satisfy
\begin{equation}
\label{eq:EO}
P(A,\hat{Y}|Y)=P(A|Y)P(\hat{Y}|Y).
\end{equation}
Returning to the example of hiring an employee, this measure implies that \textit{among the qualified applicants}, the probability of hiring two people from different demographics should be the same. That is, if two people from different demographics are both qualified, or both not qualified, the system should hire them with equal probability. Also, note that unlike demographic parity, equalized odds allows for the ideal predictor $\hat{Y}=Y$. 

In this paper, we present a new framework for designing fair predictors from data. We utilize an information theoretic approach to model the information content of variables in the system relative to one another. We use equalized odds as the criterion to assess discrimination.
In our proposed scheme, a data variable $X$, is first mapped to an auxiliary variable $U$, to decontaminate it from the discriminatory attribute as well as ensuring generalization. 
To design this auxiliary variable, for input variable $X$ and true label $Y$, we seek to find a compact representation $U$ of $X$ that contains at most a certain level of information about the variable $X$ (to avoid overfitting), but maximizes $I(Y;U)$ (quality of decision). The auxiliary variable $U$ is in turn used as the input for the prediction task.
Similar to \cite{hardt2016equality}, our framework is only based on joint statistics of the variables rather than functional forms; hence, such a formulation is more general. Furthermore, as in many cases, the functional form  of the score and underlying training data are not public. Our formulation (unlike that of \cite{hardt2016equality}, for instance) allows both $A$ and $Y$ to have arbitrary cardinality, which implies that we can have multi-level protected attributes and labels.
We cast the task of finding a fair predictor as an optimization problem and propose an iterative solution for solving this problem.
We observe that the proposed solution does not necessarily converge for some levels of fairness. This suggests that for a given requirement on the accuracy of a predictor, certain levels of fairness may not be achievable.

A somewhat similar idea to our approach is presented in \cite{zemel2013learning}, in which the authors used an intermediate representation space with elements called prototypes. However, besides the fact that in that work demographic parity is used as the measure of discrimination, the method used for choosing the prototypes is quite different. Specifically, the main approach to avoid overfitting in the learning process is limiting the number of prototypes\footnote{Unfortunately, nothing is said in that work about choosing the number of prototypes.}, while we achieve the same goal by controlling the information in the auxiliary variable about the data.
The approach in \cite{zemel2013learning} has extended in \cite{louizos2015variational} with deep variational auto-encoders with priors that encourage independence between sensitive and latent factors of variation.

The rest of the paper is organized as follows. 
In Section \ref{sec:model} we review the notion of equalized odds and introduce our model as well as the details of our proposed learning procedure.
Additionally, we propose the optimization that must be solved to address the fairness issue.
 In Section \ref{sec:solve} we propose an iterative approach for solving the optimization problem introduced. Our concluding remarks are presented in Section \ref{sec:conc}.

\section{Model Description}
\label{sec:model}

We consider a purely observational setting in which we train a predictor from labeled data.
For each sample, we have a set of attributes, which includes protected attributes such as gender, race, religion, etc.
The protected attributes are denoted by $A$. We use $X$ to denote the rest of the attributes.
We denote the true label by $Y$ and the prediction of the label $Y$ by $\hat{Y}$.
For instance, for the example regarding risk of recidivism explained in Section \ref{sec:intro}, $A$ represents the race of each individual, $X$ represents other features of that individual (which could be correlated to the individual's race) and $Y$ determines whether he/she has committed any crimes after being released from the jail.

\begin{figure}
    \centering
    \includegraphics[scale=0.55] {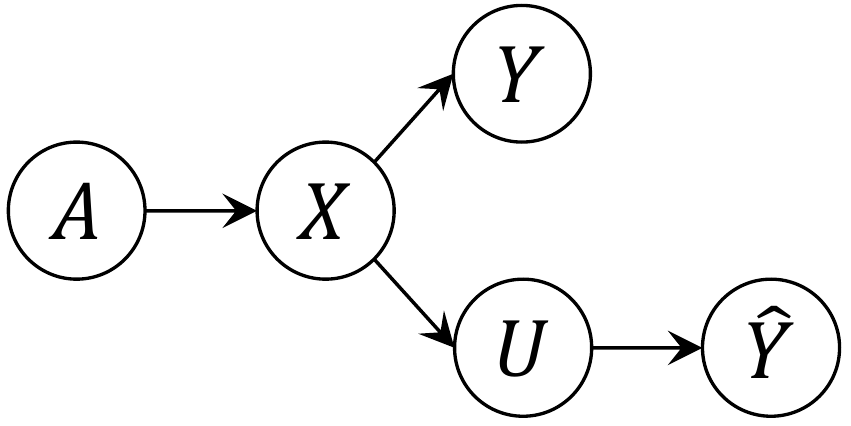}
    \caption{Graphical model of the proposed framework. $A$, $X$ and $Y$ denote the protected attribute, the rest of the attributes and the true label, respectively. $U$ is the compressed representor of $X$, which is used for designing the prediction $\hat{Y}$.}
    \label{fig:GM}
    \vspace{-3mm}
\end{figure}

The graphical model of our setup is depicted in Figure \ref{fig:GM}. As seen in this figure, $X$ and $A$ can be correlated, and given $X$, $A$ is independent of the true label $Y$. This property is essential, otherwise, the protected attribute is in fact a direct cause of the label and using this attribute in the prediction process should not be considered as discriminatory.

In order to find a fair predictor, if the joint distribution $P(A,X,Y)$ was known, we could find $P(\Hat{Y}|X)$ close to $P(Y|X)$ in the sense of equalized odds. However in reality only the empirical distribution $\Hat{P}(A,X,Y)$, which is obtained from data is available; therefore it is required to make sure that the predictor generalizes.

\textit{Generalization}: Since the number of available samples is finite, to prevent overfitting (ensuring generalization) we should constraint our hypothesis space. To do so, we compress our variable $X$ to an auxiliary variable $U$, which in turn is used for the prediction task. We also choose $U$ such that it is not contaminated by discrimination in the sense of equalized odds \cite{hardt2016equality} defined in the following.
\begin{definition}
\label{def:EO}
[Equalized odds] We say that a variable $U$ satisfies equalized odds with respect to protected attribute $A$ and outcome $Y$, if $U$ and $A$ are independent conditional on $Y$, that is,
\[
I(A;U|Y)=0.
\]
\end{definition}
\noindent
This definition is equivalent to the one in expression \eqref{eq:EO}.

Once $U$ is decontaminated from discriminatory attribute $A$, one can use any predictor to predict $Y$ from this auxiliary variable. We propose to apply a Bayesian empirical risk minimization decision rule in this work for the prediction task. 

 To obtain the mechanism for generating the auxiliary variable, we seek for a compact representation $U$ of $X$ that maximizes the utility/quality of prediction $I(Y;U)$, while it contains at most a certain level of information about the variable $X$.  
This is in essence similar to the goal in the information bottleneck (IB) method \cite{tishby2000information}.
Maximizing $I(Y;U)$ corresponds to maximizing the utility of $U$, and keeping $I(X;U)$ bounded could be viewed as regularization, which rejects complex hypotheses to ensure generalization. See \cite{xu2017information} for a detailed discussion regarding using mutual information for finding bounds on generalization error.
Note that the fact that we present fairness, accuracy and compactness via mutual information, provides us with a setting in which we do not need to have any requirement on the cardinality of variables (as opposed to \cite{hardt2016equality, zemel2013learning}).

Next, we present the details of designing the transition probability kernel for generating the auxiliary variable, as well as designing the final predictor.

\subsection{Designing the Auxiliary Variable}
\label{sec:aux}
As stated earlier, the goal of our learning scheme is to produce a compressed representor of $X$, which has as much information about the true label as possible, and is fair in the sense of Definition \ref{def:EO}. We relax the equalized odds requirement in that we allow $U$ to have a certain amount of information about the variable $A$ conditioned on $Y$. The reason for this choice will become clear in Section \ref{sec:solve}.
Therefore, the objective is to find mechanism $P(U|X)$, which maximizes $I(U;Y)$ as well as
\begin{enumerate}
\item Ensures fairness: The information shared between the protected attribute and $U$ given the true label does not exceed a certain threshold $C$, that is
\[
I(A;U|Y)\le C.
\]
\item Ensures generalization: The mutual information in $X$ and $U$ does not exceed a certain threshold $D$, that is
\[
I(X;U)\le D.
\]
\end{enumerate}
Therefore, we aim to solve the following optimization problem.
\vspace{-2mm}
\begin{align*}
&\max_{P(U|X)} ~I(U;Y)\\
&\text{s.t. }~~~I(A;U|Y)\le C,\\
&~~~~~~~I(X;U)\le D.
\end{align*}

\subsection{Designing the Predictor}
\label{sec:hat}
As stated before, after obtaining a decontaminated variable $U$, this variable can be used for the prediction task. We utilize a Bayesian decision rule described in the following.

Let $\mathcal{U}$ be the alphabet of the variable $U$ and $\mathcal{Y}$ be the alphabet of variables $Y$ and $\hat{Y}$. To quantify the quality of a decision, define a loss function $\ell:\mathcal{Y}\times\mathcal{Y}\rightarrow\mathbb{R}^+$, where $\ell(\hat{y},y)$ determines the cost of predicting $\hat{y}$ when the true label was $y$. The decisions are based on auxiliary variable $U$, which is statistically related to the true label. We denote the decision rule by $\delta:\mathcal{U}\rightarrow\mathcal{Y}$. The loss of the decision rule $\delta$ is defined as follows.
\[
L(\delta)=\mathbb{E}_{U,Y}[\ell(\delta(U),Y)].
\]
Using $L(\delta)$, the Bayesian risk minimization decision rule is 
\[
\delta^*=\arg\min_{\delta}L(\delta).
\]
For instance, for the case of binary labels with Hamming loss, defined as $\ell(y,\hat{y})=\mathbbm{1}[y\neq\hat{y}]$, we have 
\[
\delta^*(u)=\mathbbm{1}\bigg[P(Y=1|u)\ge P(Y=0|u)\bigg],
\]
which implies that we vote for the label with the maximum posterior probability.

\section{Solving the Fairness Optimization Problem}
\label{sec:solve}

In this Section, we propose a solution for the fairness optimization problem presented in Section \ref{sec:model}.
The Lagrangian for this problem will be as follows\footnote{Throughout the paper, uppercase letters for the argument of a distribution indicate all the parameters of the distribution, e.g., $P(U|X) \equiv \{P(u|x), ~\forall u,x\}$.}
\begin{equation}
\label{eq:lag}
    \mathcal{L}(P(U|X)) = \alpha I(X;U) +\beta I(A;U|Y)-I(U;Y),
\end{equation}
where the parameters $\alpha$ and $\beta$ determine the trade off between accuracy, information compression, and fairness. 

Equation \eqref{eq:lag} is similar to the objective function in \cite{chechik2003extracting}, where for given variables $X$, $Y^+$, and $Y^-$, the authors aimed to uncover structures in $P(X,Y^+)$ that do not exist in $P(X,Y^-)$, used for hierarchical text categorization.


We propose an alternating optimization method to solve the aforementioned problem. The pseudo-code of the proposed approach is presented in Algorithm \ref{alg:itt}.
In each iteration, $\mathcal{L}$ is reduced by minimizing objective function over three distributions $Q(U|X)$, $R(U)$, and $S(Y|U)$ separately. Functions $f(X,U,\alpha,\beta)$ and $Z(X,\alpha,\beta)$ are used for updating $Q(U|X)$, which are defined as follows:
\begin{equation}
\label{eq:Z}
Z(x,\alpha,\beta)=\sum_u R(u)\exp(f(x,u,\alpha,\beta)),
\vspace{-4mm}
\end{equation}
and
\begin{equation}
\label{eq:f}
\begin{aligned}
\begin{aligned}
    f&(x,u,\alpha,\beta)=\\
    &\frac{\beta}{\alpha}\sum_{y'} P(y'|x)D(P(A|x)||\frac{\sum_{x''}Q(u|x'')P(x,y,A)}{\sum_{x''}Q(u|x'')P(x,y)})\\
    &-\frac{1}{\alpha}D(P(Y|x)|S(Y|u)).
\end{aligned}
\end{aligned}
\end{equation}


\begin{theorem}
\label{thm:1}
For values of $\beta$ small enough, and any arbitrary value $\alpha$, Algorithm \ref{alg:itt} converges to a stationary point of the Lagrangian functions $\mathcal{L}$ given in equation \eqref{eq:lag}. 
\end{theorem}
\noindent
See Appendix \ref{sec:Ap1} for a proof.

\begin{algorithm}[t]
\begin{algorithmic}
\STATE {\bf Input:} Empirical distribution $\hat{P}(A,X,Y)$, initial distributions $Q^0{(U|X)}$, $R^0{(U)}$, and $S^0{(Y|U)}$ parameters $\alpha$, $\beta$, termination threshold $\epsilon > 0$. 
\STATE Initiate $\mathcal{L}^0=0$, $\mathcal{L}^1=\epsilon$, and $t=1$.
\WHILE{$\mathcal{L}^t-\mathcal{L}^{t-1}\ge\epsilon$\vspace{2mm}}
\STATE $Q^t(u|x)\leftarrow\frac{R^{t-1}(u)}{Z^{t-1}(x,\alpha,\beta)}\exp(f^{t-1}(x,u,\alpha,\beta))$, $\forall u,x$.\vspace{2mm}
\STATE $R^{t}(u)\leftarrow\sum_{x'} Q^t(u|x')P(x')$, $\forall u$.\vspace{2mm}
\STATE $S^t(y|u)\leftarrow\frac{1}{R^t(u)}\sum_{x'} Q^{t}(u|x')P(y,x')$, $\forall u,y$.\vspace{2mm}

\STATE $\mathcal{L}^{t+1} \leftarrow\alpha I(X;U) +\beta I(A;U|Y)-I(U;Y)$.\vspace{1mm}
\STATE $t=t+1$.\vspace{1mm}
\ENDWHILE
\STATE {\bf Output:} Conditional distribution $Q(U|X)$.
 \caption{Designing the conditional distribution of $U$.}
 \label{alg:itt}
\end{algorithmic}
\end{algorithm}

In general there is no guarantee that Algorithm \ref{alg:itt} converges to the global minimum of the Lagrangian. Nevertheless, experimental results show that this altenative optimization algorithm almost always converges to a local minimum of the objective function in \eqref{eq:lag}. Note that since achieving the global optimum is not guaranteed, one should initiate the algorithm from several different starting distributions.

The fact that convergence occurs only for a certain range of values for parameter $\beta$, suggests that for a given requirement on the accuracy of a predictor, certain levels of fairness may not be achievable.
This can imply an inherent bound for the level of fairness that any algorithm can achieve, a conclusion which could have not been obtained from the other existing works.
\vspace{-2mm}
\section{Conclusion}
\label{sec:conc}

We studied the problem of fairness in supervised learning, which is motivated by the fact that automated decision making systems may inherit biases related to sensitive attributes, such as gender, race, religion, etc., from the historical data that they have been trained on.
We presented a new framework for designing fair predictors from data via an information theoretic machinery.
Equalized odds was used as the criterion for discrimination, which demands that the prediction should be independent of the protected attribute conditioned on the actual label.
In our proposed scheme, a data variable is first mapped to an auxiliary variable to decontaminate it from the discriminatory attribute as well as ensuring generalization.
We modeled the task of designing the auxiliary variable as an optimization problem which aims to force the variable to be fair in the sense of equalized odds and maximizes the mutual information between the auxiliary variable and the true label, whilst keeping the information that this variable contains about the data limited.
We proposed an alternative solution for solving this optimization problem.
We observed that the proposed solution does not necessarily converge for some levels of fairness. This suggests that for a given requirement on the accuracy of a predictor, certain levels of fairness may not be achievable.
The final predictor is obtained by applying a Bayesian decision rule to the auxiliary variable.
Finding an exact bound on the achievable level of fairness, as well as applying the proposed method to real data is considered as our future work.

\appendices
\section{Proof of Theorem \ref{thm:1}} 
\label{sec:Ap1}
The Lagrangian in equation \eqref{eq:lag} can be written as follows:
\begin{equation}
\label{eq:prf2}
\begin{aligned}
\mathcal{L}(P(U|X))=&\alpha \sum_{x,u}P(x) P(u|x)\log\frac{P(u|x)}{P(u)}+\beta G(P(U|X))\\
&+\sum_{x,u,y}P(x,y)P(u|x)\log\frac{P(y|x)}{P(y|u)}-I(X;Y),
\end{aligned}
\end{equation}
where
\[
\begin{aligned}
&G(P(U|X))=I(A;U|Y)\\ 
&= \sum_{a,u,y,x} P(u|x)P(a,y,x) \log \frac{\sum_{x'}P(u|x')P(x',y,a)}{\sum_{x'}P(u|x')P(x',y)}.
\end{aligned}
\]
We note that, the only unknown parameters are $P(U|X)$, and all of the other distributions can be estimated from the given samples of $(X,Y,A)$.\\
Changing the notation of $P(u|x)$ to $Q(u|x)$ (to emphasize that it is designed), and  using \cite[Lemma 10.8.1]{cover2012elements}, we can write the optimization as follows:
\vspace{-2mm}
\[
\begin{aligned}
&\min_{Q(u|x)}\mathcal{L}(Q(U|X))=\min_{Q(u|x)}\Bigg[\alpha \sum_{x,u}P(x) Q(u|x)\log\frac{Q(u|x)}{P(u)}\\
&+\beta G(Q(U|X))
+\sum_{x,u,y}P(x,y)Q(u|x)\log\frac{P(y|x)}{P(y|u)}\Bigg] - I(X;Y)
\end{aligned}
\]
\[
\begin{aligned}
&=\min_{Q(u|x)}\Bigg[\min_{S(Y|U)}\min_{R(U)}\big[\alpha \sum_{x,u}P(x) Q(u|x)\log\frac{Q(u|x)}{R(u)}\\
&+\beta G(Q(U|X))
+\sum_{x,u,y}P(x,y)Q(u|x)\log\frac{P(y|x)}{S(y|u)}\Big]\Bigg] - I(X;Y),
\end{aligned}
\]
where the inner minimizations are over all probability distributions.
Changing the order of three minimizations, we obtain
\begin{equation}
\label{eq:finop}
\begin{aligned}
&\min_{S(Y|U)}\min_{R(U)}\min_{Q(u|x)}\alpha \sum_{x,u}P(x) Q(u|x)\log\frac{Q(u|x)}{R(u)}\\
&+\beta G(Q(U|X))
+\sum_{x,u,y}P(x,y)Q(u|x)\log\frac{P(y|x)}{S(y|u)} - I(X;Y).
\end{aligned}
\end{equation}
Since $x\mapsto x\log x$ is a convex function, and summation of a convex function with a linear function remains convex, the first and the third terms of equation \eqref{eq:finop} combined is a convex function of $Q(u|x),~ \forall u,x$. For any function $G(Q(U|X))$, there exist $\beta$ small enough such that the combination of the first three terms of equation \eqref{eq:finop} remains convex with respect to each $Q(u|x),~ \forall u,x$.

We add one more term $\lambda(x)(\sum_{u}Q({u|x})-1), ~\forall x$ to the Lagrangian for the constraint that for each $x$, $Q({u|x})$ should sum up to 1.
As a result, taking the derivative of this function with respect to $Q(u|x)~ \forall u,x$, and setting it equal to zero, the minimum of the function can be found. Below, the derivative of each term is taken separately:  

\[
L_1=\sum_{x',u'}P(x')Q(u'|x')\log\frac{Q(u'|x')}{R(u')}.
\]
Therefore,
\begin{align*}
\frac{\partial L_1}{\partial Q(u|x)}&=P(x)\log\frac{Q(u|x)}{R(u)}
+\sum_{x',u'}P(x')\times\delta_{uu'}\delta_{xx'}\\
&=P(x)\log\frac{Q(u|x)}{R(u)}+P(x).
\end{align*}

For the second term in $\mathcal{L}$ we have
\begin{align*}
L_2&=I(A;U|Y)\\
&=\sum_{a',u',y',x'} P(a',u',y',x')\log\frac{P(u'|a',y')}{P(u'|y')}.
\end{align*}
Due to the graphical model in Figure \ref{fig:GM}, we have
\[
P(a',u',y',x') = P(a')P(x'|a')Q(u'|x')P(y'|x'),
\]
Therefore,
\begin{align*}
\frac{\partial P(a',u',y',x') }{\partial Q(u|x)}
= P(a')P(x'|a')\delta_{uu'}\delta_{xx'}P(y'|x').
\end{align*}
The derivative of $P(u|a,y)$ and $P(u|y)$ can be obtained similarly. Therefore, we have

\begin{align*}
&\frac{\partial L_2}{\partial Q(u|x)} 
=\sum_{a',y'} P(y',x)P(a'|x)\log\frac{P(a'|y',u)}{P(a'|y')}\\
&=-\sum_{y'}P(y',x)D(P(A|x)||\frac{\sum_{x''}Q(u|x'')P(x,y,A)}{\sum_{x''}Q(u|x'')P(x,y)})\\
&+\sum_{y'}P(y',x)D(P(A|x)||P(A|y))
\end{align*}
For the third term in $\mathcal{L}$ we have
\[
L_3=\sum_{u',y'}P(u',y')\log\frac{S(y'|u')}{P(y')}.
\]
Therefore,
\begin{align*}
\frac{\partial L_3}{\partial P(u|x)}=&-P(x)D(P(Y|x)||S(Y|u))\\
&+P(x)D(P(Y|x)||P(Y)).
\end{align*}
Summing up all terms of the derivative and setting it equal to zero, we get the desired result in \eqref{eq:Z} and \eqref{eq:f}.

Using the calculated $Q(u|x), ~\forall u,x$, we can minimize over $R(U)$ and $S(Y|U)$. Again using \cite[Lemma 10.8.1]{cover2012elements}, minimum is achieved in marginal distributions $P(Y|U)$ and $P(U)$, which can be found from $Q(U|X)$ according to Algorithm \ref{alg:itt}. 

Regarding convergence, we note that the Lagrangian in equation \eqref{eq:lag} could be written as follows
\begin{align*}
\mathcal{L}&
= \alpha \mathbb{E}_{X}[D(P(U|x)||P(U))]\\
&+\beta \mathbb{E}_{A,Y}[D(P(U|a,y)||P(U|y))]\\
&+\mathbb{E}_{X,U}D(P(Y|x)||P(Y|u))\\
&-I(X;Y).
\end{align*}
Since the first three terms of $\mathcal{L}$ are linear combinations of KL-divergences, and hence non-negative, $\mathcal{L}$ is lower bounded by $-I(X;Y)$ which is a constant. In addition, in Algorithm \ref{alg:itt}, assuming small enough $\beta$, in each of three steps of the alternating algorithm, the value of $\mathcal{L}$ decreases. Therefore, there exists $\beta_\max$, such that for values of $\beta\le\beta_\max$, the algorithm converges to a stationary point of the objective function in \eqref{eq:lag}.

\section*{Acknowledgment}

This work was in part supported by MURI grant ARMY W911NF-15-1-0479, Navy N00014-16-1-2804 and NSF CNS 17-18952.

\bibliographystyle{IEEEtran}
\bibliography{Refs}

\begin{thebibliography}{10}
\providecommand{\url}[1]{#1}
\csname url@samestyle\endcsname
\providecommand{\newblock}{\relax}
\providecommand{\bibinfo}[2]{#2}
\providecommand{\BIBentrySTDinterwordspacing}{\spaceskip=0pt\relax}
\providecommand{\BIBentryALTinterwordstretchfactor}{4}
\providecommand{\BIBentryALTinterwordspacing}{\spaceskip=\fontdimen2\font plus
\BIBentryALTinterwordstretchfactor\fontdimen3\font minus
  \fontdimen4\font\relax}
\providecommand{\BIBforeignlanguage}[2]{{%
\expandafter\ifx\csname l@#1\endcsname\relax
\typeout{** WARNING: IEEEtran.bst: No hyphenation pattern has been}%
\typeout{** loaded for the language `#1'. Using the pattern for}%
\typeout{** the default language instead.}%
\else
\language=\csname l@#1\endcsname
\fi
#2}}
\providecommand{\BIBdecl}{\relax}
\BIBdecl

\bibitem{dwork2012fairness}
C.~Dwork, M.~Hardt, T.~Pitassi, O.~Reingold, and R.~Zemel, ``Fairness through
  awareness,'' in \emph{Proceedings of the 3rd Innovations in Theoretical
  Computer Science Conference}.\hskip 1em plus 0.5em minus 0.4em\relax ACM,
  2012, pp. 214--226.

\bibitem{hardt2016equality}
M.~Hardt, E.~Price, N.~Srebro \emph{et~al.}, ``Equality of opportunity in
  supervised learning,'' in \emph{Advances in Neural Information Processing
  Systems}, 2016, pp. 3315--3323.

\bibitem{celis2017ranking}
L.~E. Celis, D.~Straszak, and N.~K. Vishnoi, ``Ranking with fairness
  constraints,'' \emph{arXiv preprint arXiv:1704.06840}, 2017.

\bibitem{angwin2016machine}
J.~Angwin, J.~Larson, S.~Mattu, and L.~Kirchner, ``Machine bias,'' \emph{Pro
  Publica}, 2016.

\bibitem{kay2015unequal}
M.~Kay, C.~Matuszek, and S.~A. Munson, ``Unequal representation and gender
  stereotypes in image search results for occupations,'' in \emph{Proceedings
  of the 33rd Annual ACM Conference on Human Factors in Computing
  Systems}.\hskip 1em plus 0.5em minus 0.4em\relax ACM, 2015, pp. 3819--3828.

\bibitem{dwork2008differential}
C.~Dwork, ``Differential privacy: A survey of results,'' in \emph{International
  Conference on Theory and Applications of Models of Computation}.\hskip 1em
  plus 0.5em minus 0.4em\relax Springer, 2008, pp. 1--19.

\bibitem{dwork2006calibrating}
C.~Dwork, F.~McSherry, K.~Nissim, and A.~Smith, ``Calibrating noise to
  sensitivity in private data analysis,'' in \emph{TCC}, vol. 3876.\hskip 1em
  plus 0.5em minus 0.4em\relax Springer, 2006, pp. 265--284.

\bibitem{kalantari2016optimal}
K.~Kalantari, L.~Sankar, and A.~D. Sarwate, ``Optimal differential privacy
  mechanisms under hamming distortion for structured source classes,'' in
  \emph{Information Theory (ISIT), 2016 IEEE International Symposium on}.\hskip
  1em plus 0.5em minus 0.4em\relax IEEE, 2016, pp. 2069--2073.

\bibitem{zemel2013learning}
R.~Zemel, Y.~Wu, K.~Swersky, T.~Pitassi, and C.~Dwork, ``Learning fair
  representations,'' in \emph{Proceedings of the 30th International Conference
  on Machine Learning (ICML-13)}, 2013, pp. 325--333.

\bibitem{feldman2015certifying}
M.~Feldman, S.~A. Friedler, J.~Moeller, C.~Scheidegger, and
  S.~Venkatasubramanian, ``Certifying and removing disparate impact,'' in
  \emph{Proceedings of the 21th ACM SIGKDD International Conference on
  Knowledge Discovery and Data Mining}.\hskip 1em plus 0.5em minus 0.4em\relax
  ACM, 2015, pp. 259--268.

\bibitem{zafar2017fairness}
M.~B. Zafar, I.~Valera, M.~Gomez~Rodriguez, and K.~P. Gummadi, ``Fairness
  constraints: Mechanisms for fair classification,'' \emph{arXiv preprint
  arXiv:1507.05259}, 2017.

\bibitem{edwards2015censoring}
H.~Edwards and A.~Storkey, ``Censoring representations with an adversary,''
  \emph{arXiv preprint arXiv:1511.05897}, 2015.

\bibitem{louizos2015variational}
C.~Louizos, K.~Swersky, Y.~Li, M.~Welling, and R.~Zemel, ``The variational fair
  autoencoder,'' \emph{arXiv preprint arXiv:1511.00830}, 2015.

\bibitem{tishby2000information}
N.~Tishby, F.~C. Pereira, and W.~Bialek, ``The information bottleneck method,''
  in \emph{The 37th Allerton Conference on Communication, Control, and
  Computing}, 1999.

\bibitem{xu2017information}
A.~Xu and M.~Raginsky, ``Information-theoretic analysis of generalization
  capability of learning algorithms,'' in \emph{Advances in Neural Information
  Processing Systems}, 2017, pp. 2521--2530.

\bibitem{chechik2003extracting}
G.~Chechik and N.~Tishby, ``Extracting relevant structures with side
  information,'' in \emph{Advances in Neural Information Processing Systems},
  2003, pp. 881--888.

\bibitem{cover2012elements}
T.~M. Cover and J.~A. Thomas, \emph{Elements of information theory}.\hskip 1em
  plus 0.5em minus 0.4em\relax John Wiley \& Sons, 2012.

\end{thebibliography}

\end{document}